\documentclass[journal]{IEEEtran}

\usepackage{url,lineno,microtype}
\usepackage{cite}

\usepackage{hyperref}
\hypersetup{
    bookmarks=true,         
    unicode=false,          
    pdftoolbar=true,        
    pdfmenubar=true,        
    pdffitwindow=false,     
    pdftitle={TrueNorthflow},    
    pdfauthor={},     
    pdfsubject={},   
    pdfcreator={},   
    pdfproducer={},  
    pdfkeywords={}, 
    pdfnewwindow=true,      
    colorlinks=true,       
    linkcolor=red,          
    citecolor=green,        
    filecolor=magenta,      
    urlcolor=cyan           
}

\usepackage{amsmath}
\usepackage{amsfonts}
\usepackage{graphicx}
\usepackage{epstopdf}
\usepackage{algorithm}
\usepackage{algorithmic}

\usepackage{empheq}

\usepackage{float}

\usepackage[position=b]{subcaption}

\usepackage{color}

\begin{document}
\title{Spiking Optical Flow for Event-based Sensors Using IBM's TrueNorth Neurosynaptic System}

\author{Germain~Haessig, Andrew~Cassidy, Rodrigo~Alvarez, Ryad~Benosman and~Garrick~Orchard
\thanks{G. Haessig and R. Benosman are with the Vision and Natural Computation team, Institut de la vision, Universite Pierre et Marie Curie, Paris, France
  e-mail: germain.haessig@inserm.fr}
\thanks{A. Cassidy and R. Alvarez are with IBM Research, Austin, USA.}
\thanks{G. Orchard is with Temasek Laboratories and the Singapore Institute for Neurotechnology at the National University of Singapore, Singapore.}
}


\maketitle

\begin{abstract}
This paper describes a fully spike-based neural network for optical flow estimation from Dynamic Vision Sensor data. A low power embedded implementation of the method which combines the Asynchronous Time-based Image Sensor with IBM's TrueNorth Neurosynaptic System is presented. The sensor generates spikes with sub-millisecond resolution in response to scene illumination changes. These spike are processed by a spiking neural network running on TrueNorth with a 1 millisecond resolution to accurately determine the order and time difference of spikes from neighbouring pixels, and therefore infer the velocity. The spiking neural network is a variant of the Barlow Levick method for optical flow estimation. The system is evaluated on two recordings for which ground truth motion is available, and achieves an Average Endpoint Error of 11\% at an estimated power budget of under 80mW for the sensor and computation.
\end{abstract}


\IEEEpeerreviewmaketitle

\section{Introduction}
\IEEEPARstart{E}{stimation} of optical flow is a computationally intensive task. Modern artificial systems achieve ever higher accuracy on the task at the expense of more complex algorithms and more powerful computing hardware such as GPUs. On the other hand, many biological systems are able to estimate optical flow both effectively and power-efficiently. Examples include insects such as bees and drosophilia, which rely heavily on optical flow for navigation \cite{joesch2010and}, but must estimate optical flow under severe size, weight, and power constraints. While the optical flow estimated by such insect may not rival modern computer vision approaches in terms of accuracy, the estimated flow is good enough for the insects to use in navigation, and impressive considering the tight constraints under which the estimates were generated. 

In artificial systems, early optical flow estimation approaches relied on methods such as velocity-tuned filters in the frequency domain \cite{watson1983look, heeger1987model}, phase-based methods \cite{barron1994d}, gradient based methods, and correlation based methods \cite{sutton1983determination}. The accuracy of these methods for optical flow estimation on sequences of images has long since been surpassed by even more computationally intensive methods which estimate flow at multiple scales \cite{alvarez1999scale}, and more recently by convolutional neural networks \cite{dosovitskiy2015flownet}, which are mostly running on power inneficient GPUs or dedicated ASICs, which are less accurate, more general, but are more power efficient (PX4Flow \cite{honegger2013open}, optical mouse). Direction selective neurons are found as early as the retina in frogs \cite{lettvin1959frog}, cats \cite{barlow1965}, rabbits, primates, and many other animals.

A key difference between optical flow estimation in biological and artificial systems lies in the format of the data from which optical flow is extracted. Both approaches start off with light from a scene focused on an image sensor, but most artificial systems convert this light signal into a sequence of static digital images from which optical flow must later be estimated. The retina works on a different principle, transducing the light signal into a continuous electrical signal, and later into discrete spikes for communication to the lateral geniculate nucleus and downstream visual cortex.

Not all artificial methods rely on static images though. Tanner and Mead \cite{tanner1986integrated} introduced the first analog two dimensional optical flow chip, extracting a global motion from a given scene on dedicated hardware. Delbruck \cite{delbruck1993silicon} also implemented such a network in VLSI. More recently, with the new event-based cameras, some spike based techniques have been proposed \cite{benosman2012asynchronous}, but they require a dedicated computer for processing. An extensive review of motion estimation designs can be found in \cite{orchard2014bioinspired}.

Over the last decade, so-called silicon-retinae have matured to a level where they are now publicly available (described later in Section~\ref{sec:event-based_sensors}). These bio-inspired devices provide a spiking output more similar to the biological retina. Recently many papers have proposed different methods for processing data, including for estimating optical flow. Benosman~\textit{et~al.} \cite{benosman2012asynchronous} proposed a plane fitting approach which estimates the motion of sharp edges. Later, Barranco~\textit{et~al.} \cite{barranco2014contour} proposed a method which also estimates flow at edges, but their method estimates the magnitude of the spatial and temporal image gradients at the edge and then relies on a gradient-based method for optical flow estimation. Bardow~\textit{et~al.} \cite{bardow2016simultaneous} apply a variational method which simultaneously estimates both the image grayscale values and optical flow from events. By enforcing spatial and temporal smoothness constraints, their method generates optical flow estimates even for image regions where no gradient information is available.

Some optical flow estimation methods take bio-inspiration a step further and make use of biologically inspired computation using spiking neurons to extract optical flow. Examples include Brosch~\textit{et~al.} implementing a phase based method \cite{brosch2016event}, Conradt~\textit{et~al.} implementing a Reichardt detector \cite{spike-reichardt_distilled.pdf}, and Orchard~\textit{et~al.} \cite{orchard2013spiking} who simulated a method relying on synaptic delays.

In this paper we propose a Spiking Neural Network variant of the Barlow $\&$ Levick model \cite{barlow1965}, using a silicon retina coupled with IBM's TrueNorth Neurosynaptic System.


The method exploits precise spike timing provided by the silicon retina to reliably extract motion direction and amplitude from a scene in real-time. Such methods promise to allow for low power visual motion estimation in real-time. TrueNorth is estimated to consume $70$mW and the silicon retina consumes approximately $10$mW (chip only, omitting FPGA for communication, which can be removed for dedicated applications). This setup can then easily be used in embedded devices such as drones or autonomous driving.\\
Section~\ref{sec:background} introduces both the model, IBM's TrueNorth Neurosynaptic System, and the silicon retina. Details of the model implementation are given in Section~\ref{sec::implementation}. Section~\ref{sec:testing} describes how the model was tested, and results of testing are described in Section~\ref{sec:results}, followed by discussion in Section~\ref{sec:discussion}.

\begin{figure}
\centering
\includegraphics[width=0.5\textwidth]{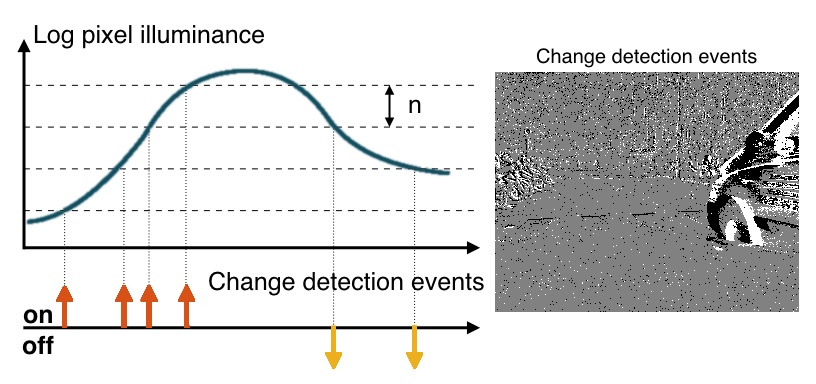}
\caption{Working principle of the ATIS camera. (left) A significant relative change of the luminance (programmable threshold $n$) generates an ON (respectively OFF) event when there is an increase (resp. decrease) in luminosity. (right) Snapshot of events over $100$ ms for a car passing in front of the camera. White dots represents ON events, black ones OFF events.}
\label{fig::atis}
\end{figure}

\section{Background}\label{sec:background}

\subsection{Direction Sensitive (DS) Unit}
\begin{figure}
\centering
\includegraphics[width=0.45\textwidth]{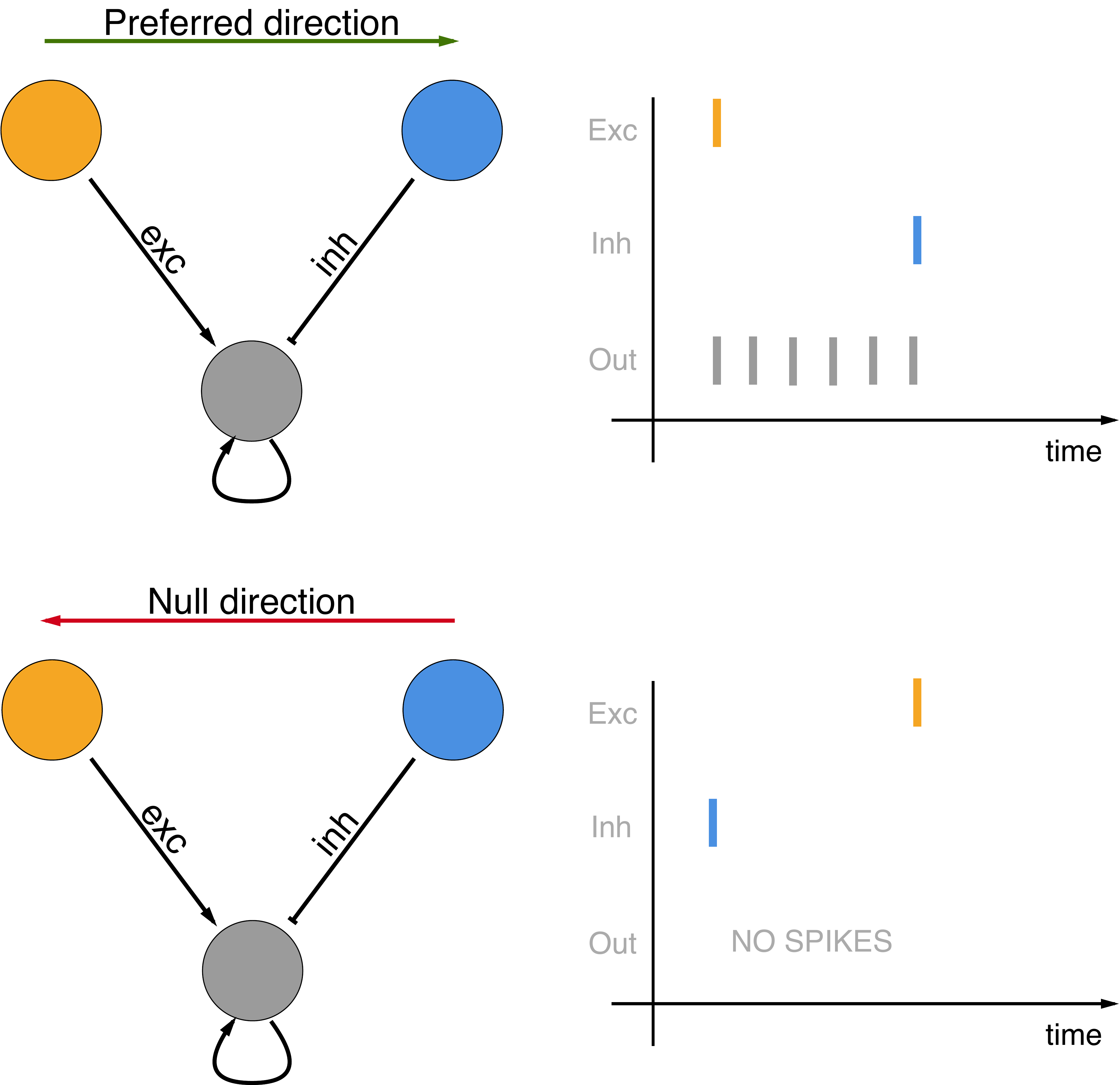}
\caption{Direction Sensitive (DS) unit. A stimulus from left to right first excite (yellow cell) the output (gray), which is self-excitatory until it gets an strong inhibitory input (blue cell), as a movement in the other direction (right to left) doesn't produce any output. Thus, this unit is directive to a specified direction. More, the timing between the beginning and the end of the output spiking give an information of the (inverse) velocity. Inspired by \cite{giulioni2016event}.}
\label{fig::flow1}
\end{figure}

The basis of the network, introduced in \cite{giulioni2016event}, is described in Fig. \ref{fig::flow1}. The original model, inspired by the neural circuitry found in the rabbit's retina by Barlow and Levick \cite{barlow1965} is based on inhibition-based direction-sensitive units, combined in motion detectors.

Fig.~\ref{fig::flow1} shows a simplified Direction Selective (DS) neuron which responds (spikes) to the detection of motion in its preferred direction (rightward). The DS neuron (gray) receives inputs from two neighbouring ATIS pixels (bipolar cells). Each ATIS pixel (orange and blue) will generate an output spike when stimulated by the passing of an edge. For motion in the preferred direction (rightward), the edge will pass over the orange pixel first, and the blue pixel second.

When the edge passes over the orange pixel, a spike is generated and the DS neuron is excited. The excitatory input triggers a burst of spikes from the DS neuron. The burst continues until the edge passes over the blue pixel, at which point the blue pixel will generate an output spike which inhibits the DS neuron, causing the spike burst to end. The time-length of the spike burst encodes the time taken for the edge to pass from the orange pixel to the blue pixel, thus providing information on the edge velocity in the preferred direction.

For motion in the opposite direction (leftward) the inhibition from the blue pixel will arrive before the excitation from the orange pixel. Due to the initial inhibition, the later excitation from the orange pixel is not sufficient to drive the DS neuron membrane potential above threshold, and thus the DS neuron does not spike in response to leftward motion.

\begin{figure}
\centering
\includegraphics[width=0.45\textwidth]{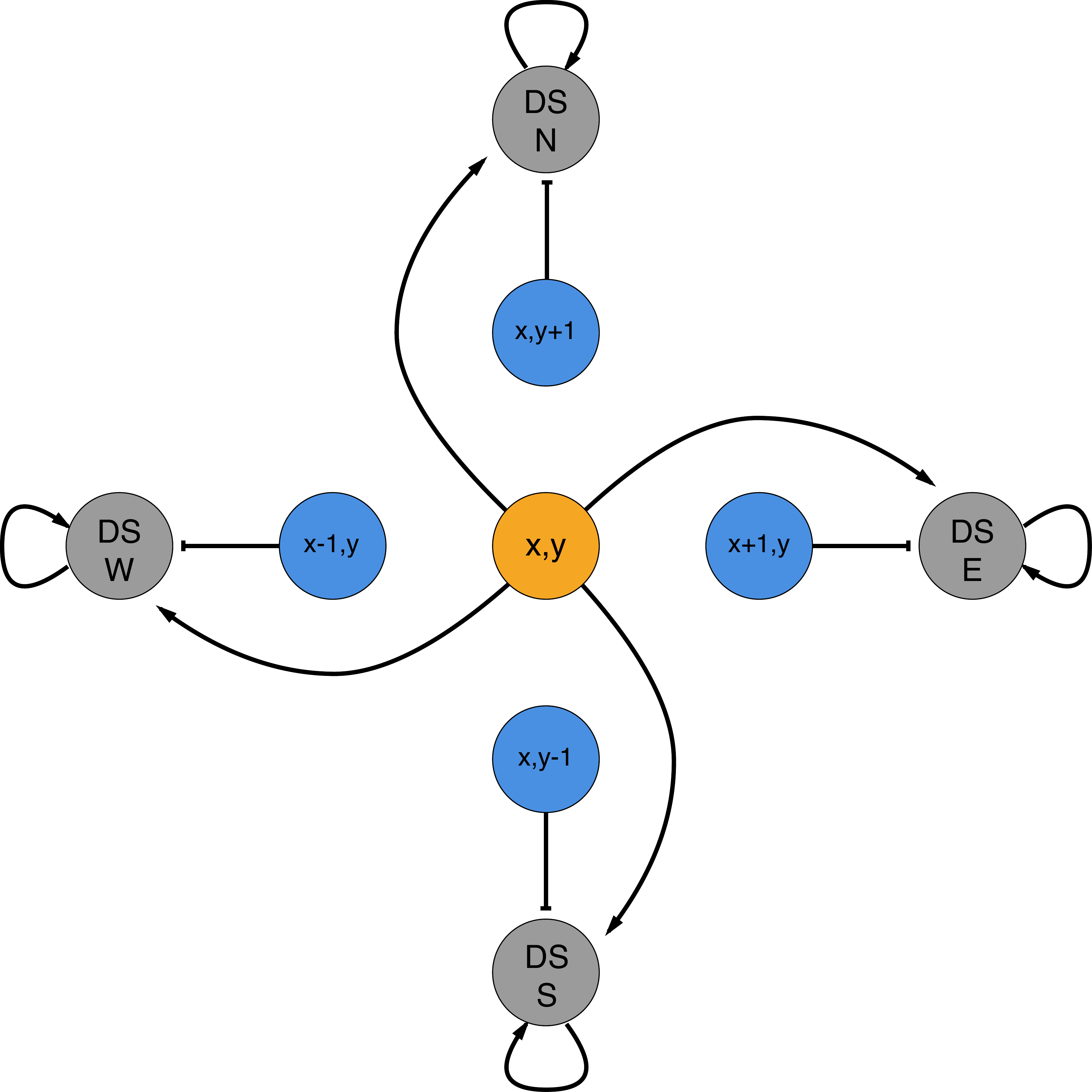}
\caption{Assembly of the 4 DS units (North, South, West, East). The direction of the movement is gathered by taking in consideration, for each pixel, a vertical and horizontal component. Inspired by \cite{giulioni2016event}.}
\label{fig::flow2}
\end{figure}

The full direction is obtained with a combination of four single DS units, as shown in Fig.~\ref{fig::flow2}.

In implementation, there is a possibility of receiving an isolated noise spike from the orange pixel, which would cause the DS neuron to begin bursting, and continue bursting indefinitely. To handle this case, a delayed copy of the spikes from the orange pixel are used to inhibit the DS neuron, thus limiting the maximum burst length to the length of delay used.

\subsection{Event-based Sensor}\label{sec:event-based_sensors}
Conventional image sensors sample the visual scene at fixed temporal periods (framerate). All pixels acquire light in a synchronous fashion by integrating photons over a fixed time-period. When observing a dynamic scene, this framerate, no matter its value, will always be wrong because there is no relation whatsoever between the temporal dynamics of the scene and the acquisition period. This leads to simultaneous over-sampling and under-sampling of different parts of the scene, motion blur for moving objects, and data redundancy for static background.

Event-based sensors \cite{posch2014retinomorphic}, also known as silicon retinae, are an alternative to fixed-frequency imaging devices. In these devices, a time-varying signal is sampled on its amplitude-axis instead of time-axis, leading to a non-uniform sampling rate that matches the temporal dynamics of the signal, as shown on Fig.~\ref{fig::atis}. The Asynchronous Time-based Image Sensor (ATIS\cite{posch11}) is an QVGA array of asynchronous independent pixels, each combining a change detection unit and an absolute gray level measurement unit. Each time a pixel detects a significant change in luminance in its field of view, an event is generated. Each event can be represented as a triplet $ev_i = \left(\textbf{x}_i,t_i,p_i\right)$, $ev_i$ being the $i-th$ event, where $\textbf{x}_i$ is the event spatial coordinates, $t_i$ its timestamp and $p_i$ its polarity, indicating if the change is an increase or decrease in luminance. In this work, only the change detection unit is used.

\subsection{The TrueNorth Environment}
A TrueNorth Chip \cite{merolla2014million} consists of a network of 64$\times$64=4096 neurosynaptic cores with programmable connectivity, synapses, and neuron parameters. Connectivity between neurons follows a block-wise scheme. Each core has 256 input axons, with programmable connectivity to any of the 256 neurons in that core. Each neuron's output can be routed to a single axon anywhere on the chip. All communications to, from, and within chip are performed asynchronously \cite{merolla2016deep}.

\begin{figure}[t!]
    \centering
    \begin{subfigure}[c]{0.22\textwidth}
        \centering
        \includegraphics[width=\textwidth]{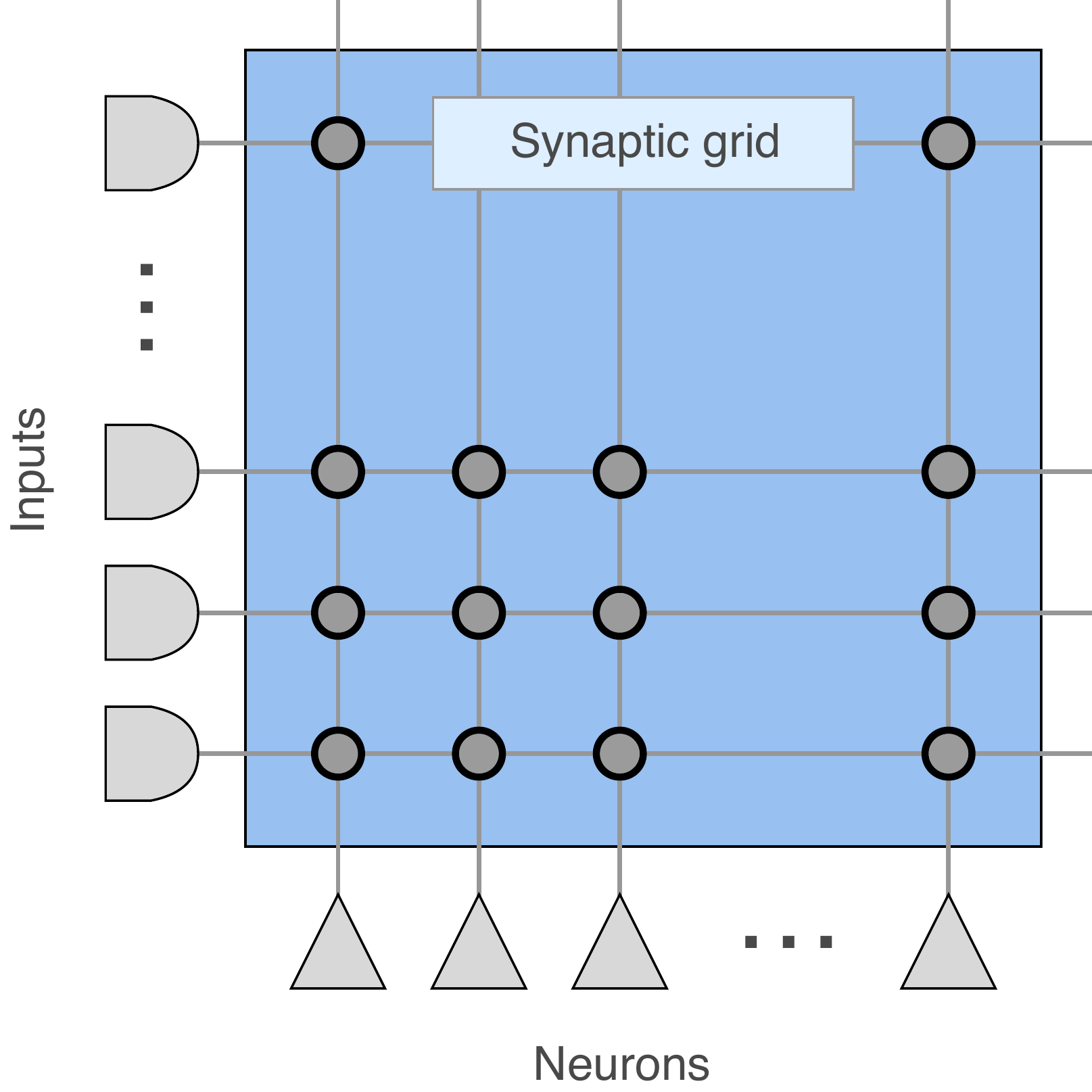}
        \caption{}
    \end{subfigure}%
    ~
    \begin{subfigure}[c]{0.22\textwidth}
        \centering

        \includegraphics[width=\textwidth]{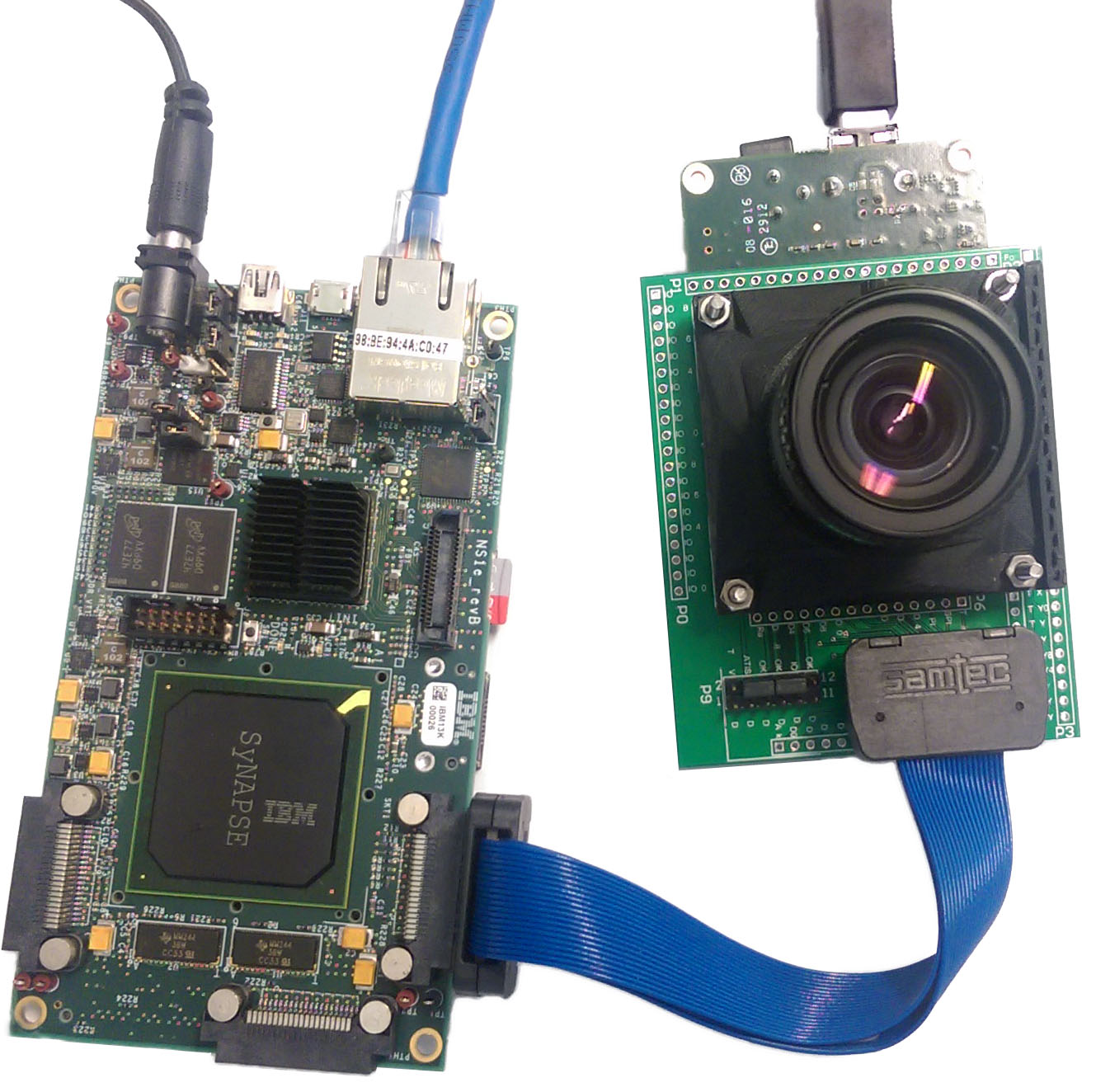}

        \caption{}
    \end{subfigure}
    \caption{a) The TrueNorth topology. Each neurosynaptic core has 256 axons, 256$\times$256 synapse crossbar and 256 neurons. Information flows from axons to neurons gated by binary synapses, where each axon fans out, in parallel, to all neurons thus achieving a 256-fold
reduction in communication volume compared to a point-to-point approach. Network operation is governed by a discrete time step. In a time
step, if the synapse value for a particular axon-neuron pair is non-zero and
the axon is active, then the neuron updates its state by the synaptic weight
corresponding to the axon type. Next, each neuron applies a leak, and any
neuron whose state exceeds its threshold fires a spike \cite{cassidy2013cognitive}. b) IBM's NS1e board (4096 cores, 1 Million neurons and 256 Million synapses) and ATIS sensor with native link used for this paper.}
\label{fig::TNcore}
\end{figure}

Each TrueNorth neurosynaptic core is made of 256 axons, 256$\times$256 synapse crossbar and 256 neurons (Fig.~\ref{fig::TNcore} A). In this paper, the NS1e board was used, containing 4096 cores, $1M^+$ neurons, $268M^+$ synapses, embedded in a board with a Zynq FPGA containing an ARM-core (Fig.~\ref{fig::TNcore} B).

\section{Implementation}
\label{sec::implementation}
TrueNorth implementation of the model was achieved by carefully configuring a single tile-able TrueNorth core, thanks to the tools presented in \cite{amir2013cognitive}, to handle a small local region of the input space, and then tiling enough copies of these cores to cover the full input image.

\begin{figure}
\centering
\includegraphics[width=0.5\textwidth]{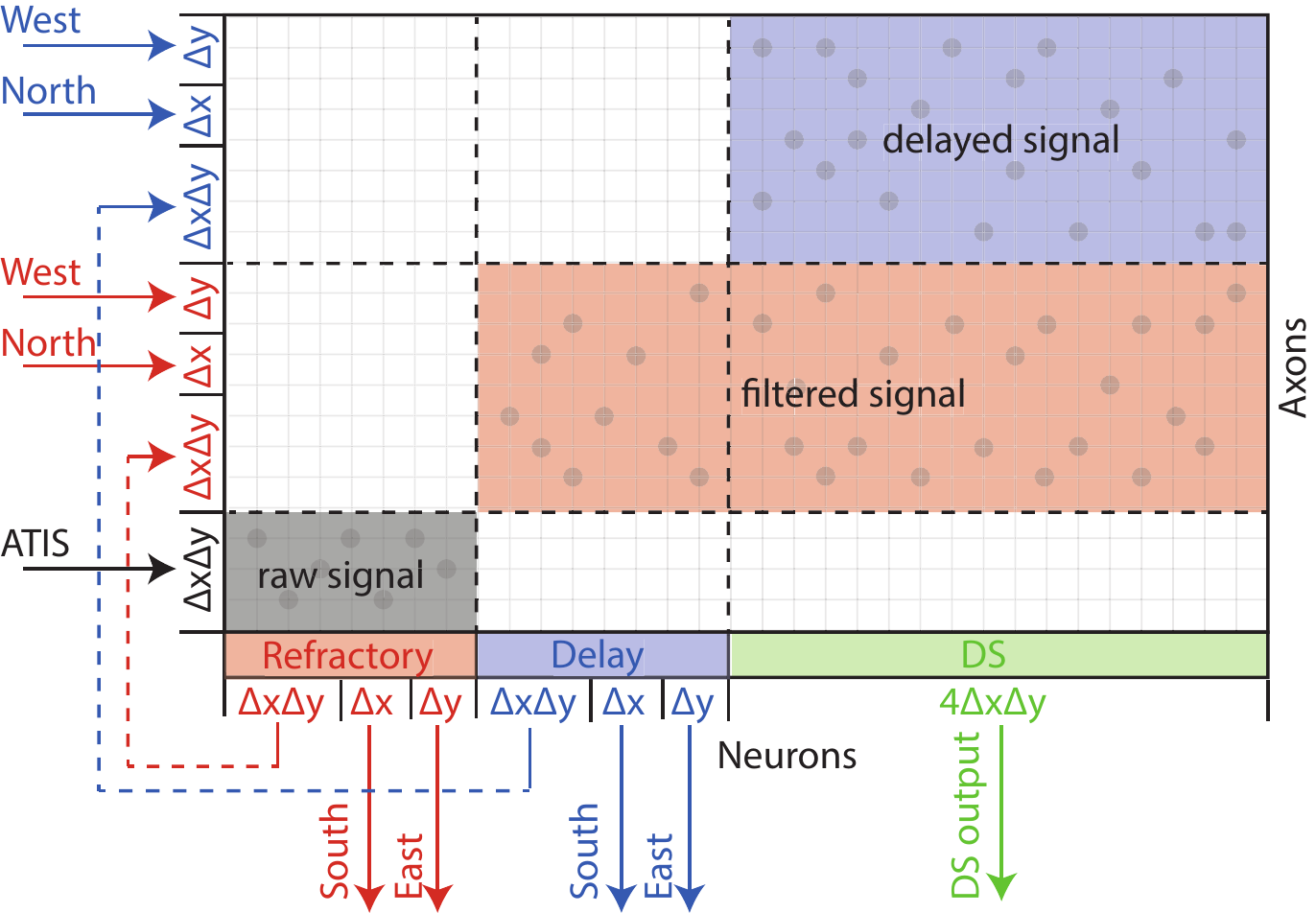}
\caption{Arrangement of a TrueNorth core implementing the motion model. Refractory (red), Delay (Blue), and DS (green) neurons are shown horizontally along the bottom of the core. Input axons are shown vertically on the left of the core. Text labels indicate the direction inputs are arriving from and outputs are routed to. The number of neurons (horizontal axis) and axons (vertical axis) are explicitly indicated for each part of the model. Shaded regions indicate areas where connections (synapses) exist between axons on the left and neurons at the bottom.}
\label{fig::core}
\end{figure}

Within each core, neurons are divided into three main modules, shown in different colors in Fig.
~\ref{fig::core}. The first module (red) receives input spikes from ATIS and generates a copy of these spikes which can be used for further processing by the other two modules. The second module (blue) implements the delay required to generate the delayed inhibition for the DS neurons. The third module (green) implements the DS neurons. Each module is described further below.

\subsection{Input Module}
The input stage provides a copy of the input spikes which serve as inputs to the DS neuron module and delay module. The input stage is also used to enforce a refractory period which limits the minimum allowable time between two input spikes from the same pixel by blocking the second spike if it falls within $\tau_{r}$ of the first.

A refractory period of $\tau_{r}$ (ms) is implemented using a \textit{refractory} neuron model (see Table~\ref{tab:neuron_params}). When at rest state (membrane potential of zero), the neuron should fire an output spike as soon as an input spike is received. This is achieved by enforcing
\begin{equation}\label{eq:thresh}
\begin{array}{l l}
w_e + l \geq \Theta \\
\end{array}
\end{equation}
where $w_e$ is the excitatory synaptic weight, $l$ is the leak, and $\Theta$ is the neuron threshold.

After spiking, the neuron resets to a large negative voltage and slowly leaks back to zero, thus implementing a refractory period. The desired refractory period is achieved by setting the reset and leak values such that
\begin{equation}
\begin{array}{l l l}
\tau_r &\approx \frac{V_r}{l} \\
V_r &= -(2^n -1), & n\in \mathbb{N}\\
\end{array}
\end{equation}
where $\tau_r$ is the desired refractory period, and $V_r$ is the reset potential, which can only take on limited values due to constraints of the chip.

Input spikes received while the neuron potential is leaking towards zero will not cause an output spike, but they will shorten the refractory period. The actual refractory period achieved, $\tau'_{r}$, is given by:
\begin{equation}
\begin{array}{l l}
\tau'_{r} &= \frac{V_r + w_es}{l}\\
\\
          &= \tau_r + \frac{w_es}{l}\\
\end{array}
\end{equation}
where $s$ is the number of spikes received during the refractory period $\tau'_{r}$. To ensure that the achieved refractory period $\tau'_{r}$ is as close as possible to the desired refractory period, $\tau_{r}$, the term $w_es/l$ must be minimized. The minimum value which still satisfies constraint \eqref{eq:thresh} occurs when $l=254$, $w_e=255$.

Connectivity on TrueNorth is bound by certain constraints, one of which is that a spike can only be routed to one core. This constraint can be overcome by implementing multiple identical neurons on the same core. Each neuron will then generate an identical output, thereby providing multiple copies, each of which can be routed to different cores.

In the case of our TrueNorth model, there are four co-located DS neurons at every pixel location, and they each require input from one of the four neighbouring pixels. Thus, the spikes from some pixels serve as input to neighbouring DS neurons which reside on different cores. The input stage takes care of generating copies of the spikes from these pixels and routing them to the neighbouring cores.

Each core only sends spikes to its neighbouring cores lying to the South and East, and each core receives input spikes from its neighbouring cores in the North and West.

The number of neurons required by the input module is therefore
\begin{equation}
N_{in}(\Delta x, \Delta y) = \Delta x \Delta y + \Delta x + \Delta y,
\end{equation}
where $\Delta x$ and $\Delta y$ specify the pixel size of the image region in $x$ and $y$ direction processed by a single core.

The input module implementing refraction is shown in red in Fig.~\ref{fig::core}.

\subsection{Delay Module}
The delay module implements the delay required by the delayed DS neuron inhibitory input. Delays of $\tau_d$ milliseconds are implemented using the \textit{Delay} neuron model of Table~\ref{tab:neuron_params}.

When at rest (zero membrane potential), then neuron will remain at rest until an input spike is received. An input spike will push the membrane potential above zero, and once above zero the neuron membrane potential will leak towards a positive threshold. The delay is implemented by the time taken to leak to the positive threshold. Once the threshold is reached, an output spike is emitted, and the neuron will be reset back to the rest state.

A disadvantage of the second approach is that if two input spikes arrive from the same pixel at times $t_1$ and $t_2$ such that $t_2-t_1 < \tau_d$, only a single delayed output spike will be generated, and the output spike will be generated with delay $\tau_d-1$ millseconds after $t_1$. To avoid such complications, we use the refractory period from the input stage to limit the probability of having two spikes occur close together. More specifically, we choose $\tau_r$ such that $\tau_r>\tau_d$.

All input spikes from the region covered by the core must be delayed, and copies of the delayed spikes at the South and East boundaries must be generated for neighbouring cores. Thus the total number of neurons required for the delay module, $N_{delay}(\Delta x, \Delta y)$, is
\begin{equation}\label{eq:delay_count}
\begin{array}{l l}
N_{delay}(\Delta x, \Delta y) &= N_{in}(\Delta x, \Delta y) \\
&= \Delta x \Delta y + \Delta x + \Delta y.
\end{array}
\end{equation}

The delay module (blue in Fig.~\ref{fig::core}) receives inputs from the refractory input module (red), as well as input from the refractory input modules of the cores to the North and West. The delay module generates outputs which feed back to the same core, as well as output for use in processing of the cores to the South and East.

%

%
%

\subsection{DS Module}
The DS module holds the DS neurons. The parameters for each neuron are shown in Table~\ref{tab:neuron_params}. There are four DS neurons per image location, so a core covering an image region of size $\Delta x$ by $\Delta y$ pixels will require
\begin{equation}
N_{DS}(\Delta x, \Delta y) = 4\Delta x\Delta y
\end{equation}
neurons.

The DS neurons are shown in green in Fig.~\ref{fig::core} and receive input from both the input refractory module, and the delay module of the current core, as well as from the cores to the North and West.


\subsection{Parameters}
TrueNorth neurons are a variant of linear leaky integrate and fire model with 23 tunable parameters, as mentioned in \cite{cassidy2013cognitive}. Table \ref{tab:neuron_params} contains the neuron parameters for each population.

\begin{table}[!t]
\renewcommand{\arraystretch}{1.3}
\caption{Neuron parameters}
\centering
\begin{tabular}{|c|c|c|c|c|}
\hline
Parameter                      & Symbol & Refractory         & DS neuron & Delay\\
\hline
excitatory weight (mV)              & $w_e$      & $255$              & $150$       & $1$\\ \hline
inhibitory weight (mV)              & $w_i$      & $0$                & $50$        & $0$\\ \hline
threshold (mV)                      & $\Theta$   & $1$              & $125 $      & $\tau_d$\\ \hline
leak (mV/ms)                        & $l$        & $-254$             & $-1$        & $1$\\ \hline
reset potential (mV)                & $V_r$      & $-\tau_{r}$        & $127$       & $0$\\ \hline
negative floor (mV)                 & $\beta$    & $-\tau_{r}$        & $-50$       & $0$\\ \hline
\end{tabular}
\label{tab:neuron_params}

\end{table}

The total number of neurons $N_{\sum}$ required by a core is
\begin{equation}
\begin{array}{l l l}
N_{\sum}(\Delta x, \Delta y) &= &N_{in}(\Delta x, \Delta y) + N_{delay}(\Delta x, \Delta y)  \\
&  &+ N_{DS}(\Delta x, \Delta y) \\
\\
&= &6 \Delta x \Delta y + 2 \Delta x + 2 \Delta y\\
\\
N_{\sum}(\Delta x, \Delta y) &\leq &256.\\
\end{array}
\end{equation}

The values $\Delta X = 6$ and $\Delta Y = 6$ maximize the image area processed by the core, subject to the constraint that the core only has 256 neurons. The number of axons used is less than the number of neurons, so neurons are the limiting factor.

285 identity cores are used to relay the ATIS input to the motion model (see Section~\ref{sec::atisTNlink}). For the motion model to cover the 304$\times$240 pixel input image requires another $304/6\times240/6\approx 51\times40 = 2040$ cores. In total the TrueNorth model uses $2040+285=2325$ cores to compute motion at all locations in the input image space. 240 neurons are used per core (a usage of 256 neurons per core is achievable if $\Delta X = 18$ and $\Delta Y = 2$, but each core would still only be processing 36 pixels, and thus such an arrangement is less efficient).

\subsection{Interpreting the Result}
To interpret the output spikes from the network at a particular pixel location, the outputs of all four of the DS neurons located at the pixel are combined. The same input spike will excite all four DS neurons, so they should all start bursting simultaneously (unless they are already inhibited). We use the notation $t_{x^+}$ to denote the length of the burst from the neuron sensitive to motion in the positive x-direction. The velocity can then be calculated using
\begin{equation}
\begin{array}{l l l}
t_x &= &t_{x^+} - t_{x^-}\\
t_y &= &t_{y^+} - t_{y^-}\\
v_x &=& t_x/(t_x^2+t_y^2)\\
v_y &=& t_y/(t_x^2+t_y^2)\\
\end{array}
\end{equation}
where $v_x$ and $v_y$ indicate the component of velocity in the $x$ and $y$ directions respectively, in units of pixels per millisecond.

\subsection{Self excitatory state}
The output neuron of the DS unit should be self-excitatory until it receives a strong inhibitory spike. This behavior was achieved by setting the reset potential to be higher than the threshold ($\alpha_j$). Then, when the membrane potential crosses once the threshold, the cell enters a self-excitatory state until a strong inhibition link is received.

Sometimes, due to noise, the inhibitory spike can be missing. To prevent a cell to spike indefinitely, and second strong inhibition spike is generated and delayed (typically 100ms) as the excitation spike is generated.

\subsection{ATIS-TrueNorth link}
\label{sec::atisTNlink}
The ATIS and TrueNorth chips both use variants of the Address Event-Representation (AER) protocol \cite{Boahen2000}. However, the two are not directly compatible, so the Opal Kelly XEM6010 board with Xilinx Spartan 6 FPGA which powers and configures ATIS is also used to convert the ATIS AER signals to TrueNorth AER signals.

The ATIS AER data consists of an absolute pixel address and a polarity, all of which are communicated in parallel when the AER request line is active.

The TrueNorth AER data consists of a relative core address, an axon address, and a 4 bit target time. The data is communicated in two phases, with half the bits being communicated in parallel during each phase.

The TrueNorth reset and 1kHz clock tick signals are shared with the Opal Kelly board so that it can keep track of the state of TrueNorth's internal 4 bit time. All events are communicated to TrueNorth with a target time 2ms later than the current time.

A one to one mapping from ATIS pixels to TrueNorth neurons is generated such that each TrueNorth core accepts events from a 16$\times$16 pixel region, with polarity being ignored. An array of 19$\times$15 = 285 identity cores is instantiated at the physical core locations targetted by the ATIS interface. These identity cores generate a copy of the spikes they receive, which can then be routed to the rest of the TrueNorth model.

This interface arrangement uses an extra 285 cores, but allows the model to be rapidly changed by only reconfiguring TrueNorth, instead of having to reprogram the ATIS FPGA to target different cores and neurons for different TrueNorth models.

\section{Testing}\label{sec:testing}

\subsection{Sources of Visual Data}
\label{sec:model_data}
\begin{figure}
\centering
  \includegraphics[width=0.97\columnwidth]{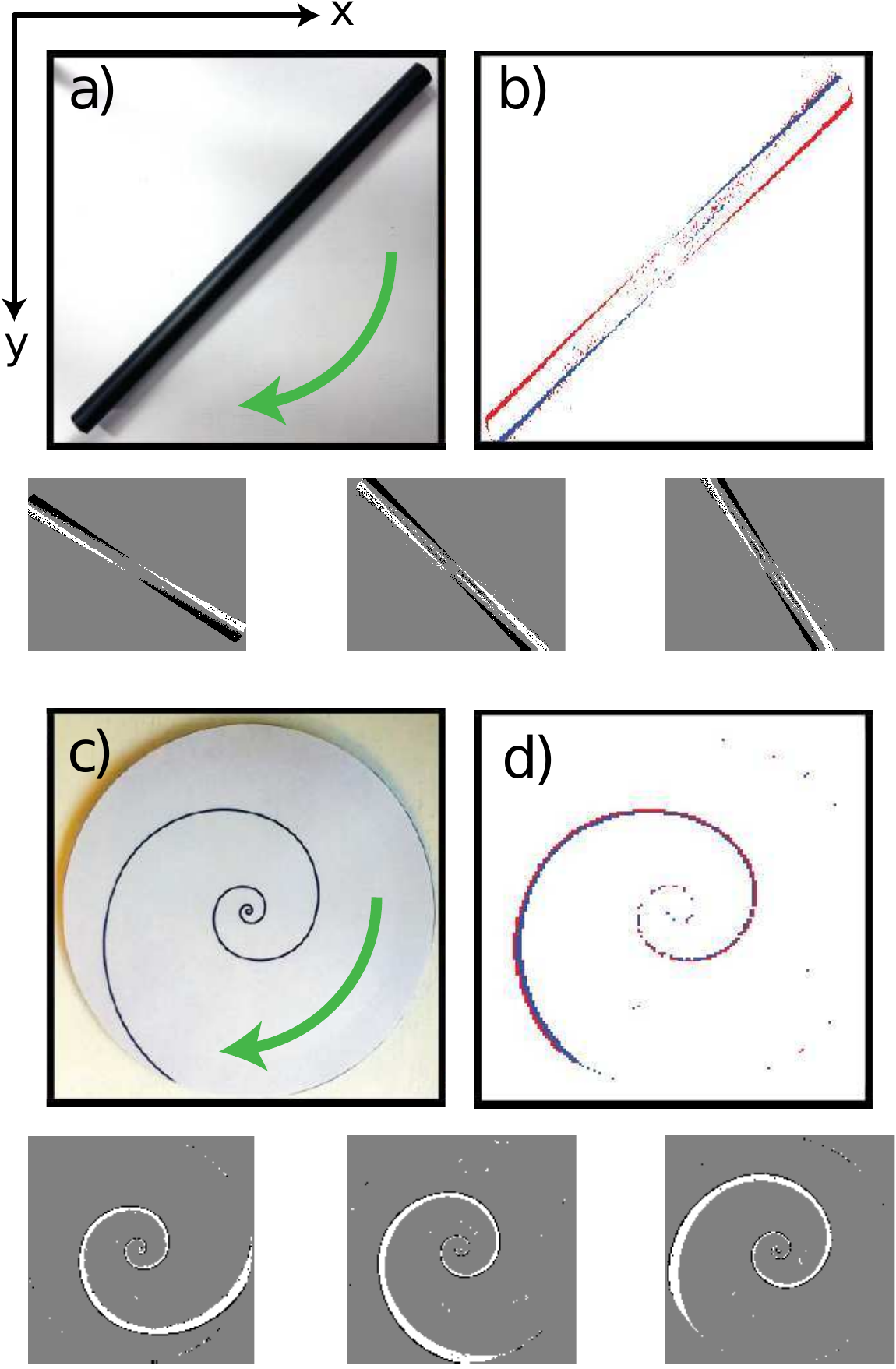}
    \caption{The two recordings used in this paper. (a) and (b) show images of a rotating pipe and spiral respectively from the sensor's viewpoint. The direction of rotation is shown by the green arrows inset. (c) and (d) show 10ms each of the pipe and spiral recordings. Red points indicate OFF-events (decreases in intensity) while blue points indicate ON-events (increases in intensity). Axes directions used in \eqref{eq:pipe} and \eqref{eq:spiral} are shown in the top left.}\label{fig:VisualSources}
\end{figure}

Two sources of visual data are used in this work: a recording of a black pipe rotating in front of a white background, originally presented in \cite{orchard2013spiking}, and a recording of a rotating spiral \cite{orchard2014bioinspired}. The pipe and spiral sequences are both ATIS recordings in which ground truth for motion can be predicted to quantify the accuracy of the output. The pipe and spiral recordings are shown in Fig.~\ref{fig:VisualSources}. This motion is modelled as described below.

\subsection{Modelling Motion for the Rotating Pipe}
We use a recording of just over half rotation of the pipe due to symmetry (the second half of the rotation will look almost identical to the first). A full rotation takes roughly 2.85 seconds, so we use a 1.5 second recording. The rotating pipe was modelled as having two parallel edges spaced a width of $w$ apart from each other and rotating about a point centered between them. Motion of the pipe could then be modelled using
\begin{equation}
\begin{array}{l l}
R(t) &= \left[\begin{matrix}\cos{(t\omega_t)} &  -\sin{(t\omega_t)}\\ \sin{(t\omega_t)}  & \cos{(t\omega_t)}\end{matrix}\right] \\
\\
Location(l,t) &= \left[\begin{matrix}x_c\\ y_c\end{matrix}\right] + R(t)\left[\begin{matrix} \pm \frac{w}{2} \\ l\end{matrix}\right]\\
\\
Direction(t,l) &=  \left\{\begin{matrix} t\omega_t & l \leq 0 \\ t\omega_t+\pi & l > 0 \end{matrix}\right. \\
\\
Speed(l) &= |l|\omega_t\\
\\
l &\in [-150,150]\\
 t &\in [0,\frac{\pi}{|\omega_t|}]\\
\omega_t = 2.21 rad/s\\
\end{array}\label{eq:pipe}
\end{equation}
where $R(t)$ is a rotation matrix which varies with time $(t)$ to model rotation of the pipe, $\omega_t$ is the angular velocity of the pipe, $Location(l,t)$ gives the pixel location, $[x,y]^T$, of points on the pipe as a function of time, their position on the edge of the pipe $l$, and the location of the center of the pipe $[x_c,y_c]^T$. The speed of each point is dependent only on its position on the pipe $l$, while the direction depends on $t$ and $l$ because one half of the pipe is moving in the opposite direction to the other. The pipe has a finite length of 300 pixels, thereby limiting $l$. The recording is long enough for half a rotation of the pipe, thereby limiting $t$. The speed and direction given by \eqref{eq:pipe} are the components perpendicular to the edges of the pipe. The direction of the $x$ and $y$ axes are shown top left of Fig.~\ref{fig:VisualSources}.

\subsection{Modelling Motion for the Rotating Spiral}
The spiral stimulus does not exhibit the same symmetry as the pipe, so we use a recording of a full rotation of the spiral, which takes 0.5 seconds. The shape of the spiral stimulus was parameterized by angle with a variable $\theta_0$ from which the speed, direction, and location of motion can be modelled using
\begin{equation}
\begin{array}{l l}
r(\theta_0)                       &= 2^{\frac{\theta_0}{\pi}}\\
\\
Location(\theta_0,t)              &= \left[\begin{matrix}x_c\\ y_c\end{matrix}\right] + \left[\begin{matrix}r(\theta_0)\cos{(-\theta_0+t\omega_t)}\\
                                                                                                            r(\theta_0)\sin{(-\theta_0+t\omega_t)}\end{matrix}\right] \\
\\
Speed(\theta_0,t)                 &= 2^{\frac{\theta_0}{\pi}}\frac{\ln{2}}{\pi}\omega_t\frac{1}{\cos{(\frac{\ln{2}}{\pi\omega_t})}}\\
\\
Direction(\theta_0,t)             &= -\theta_0 + t\omega_t + \sin{(\frac{\ln{2}}{\pi\omega_t})}\\ 
\\
\theta_0 &\in [0,20]    \\
t &\in [0,\frac{\pi}{|\omega_t}|]\\
\omega_t &= -12.57 rad/s\\
\end{array}\label{eq:spiral}
\end{equation}

where $r(\theta_0)$ is the radial distance to the spiral from the center, $[x_0, y_0]^T$. $Location(\theta_0,t)$ describes the $[x,y]^T$ location of the spiral edges. $Speed(\theta_0,t)$ and $Direction(\theta_0,t)$ give the speed and direction of motion respectively. The Cosine and Sine terms in the speed and direction equations account for the fact that the spiral is not perpendicular to the radial vector (second term in $Location(\theta_0,t)$).

\section{Results} \label{sec:results}
Output spikes are sent from the board as UDP packets. A simple UDP receiver code for visualizing the output spikes was developed that captures UDP packets, decodes them and plots the results of the flow (direction and velocity).

\begin{figure}
\centering
\includegraphics[width=0.49\textwidth]{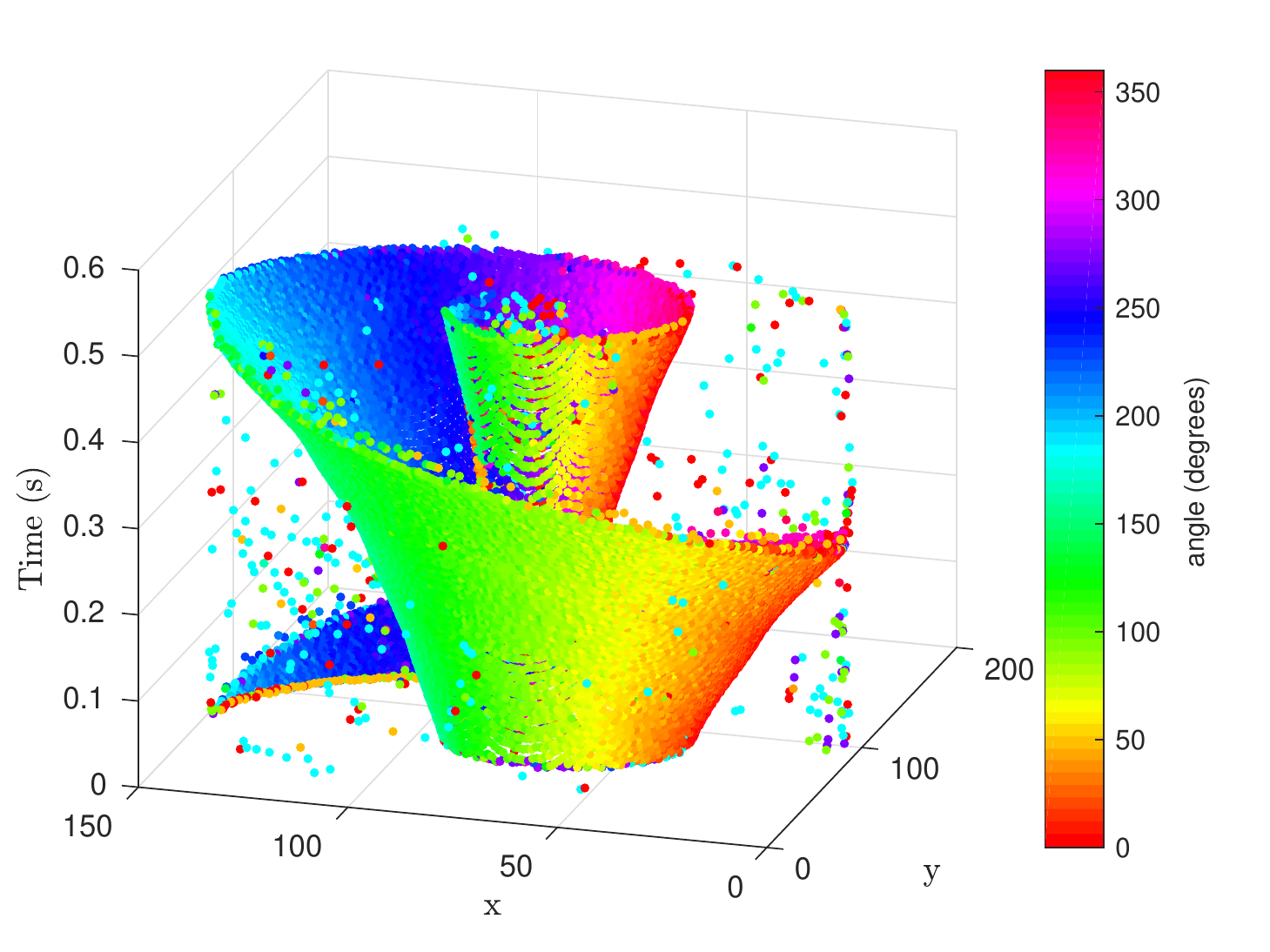}
\caption{3D representation of network result on the Spiral data. The direction of movement is color-coded in the HSV space. Some individual isolated spikes are flow estimates generated in response to sensor noise. Video available online \cite{spiral_video}.}
\label{fig::flow_out1}
\end{figure}

\begin{figure}
\centering
\begin{subfigure}[b]{0.49\textwidth}
\includegraphics[width=\textwidth]{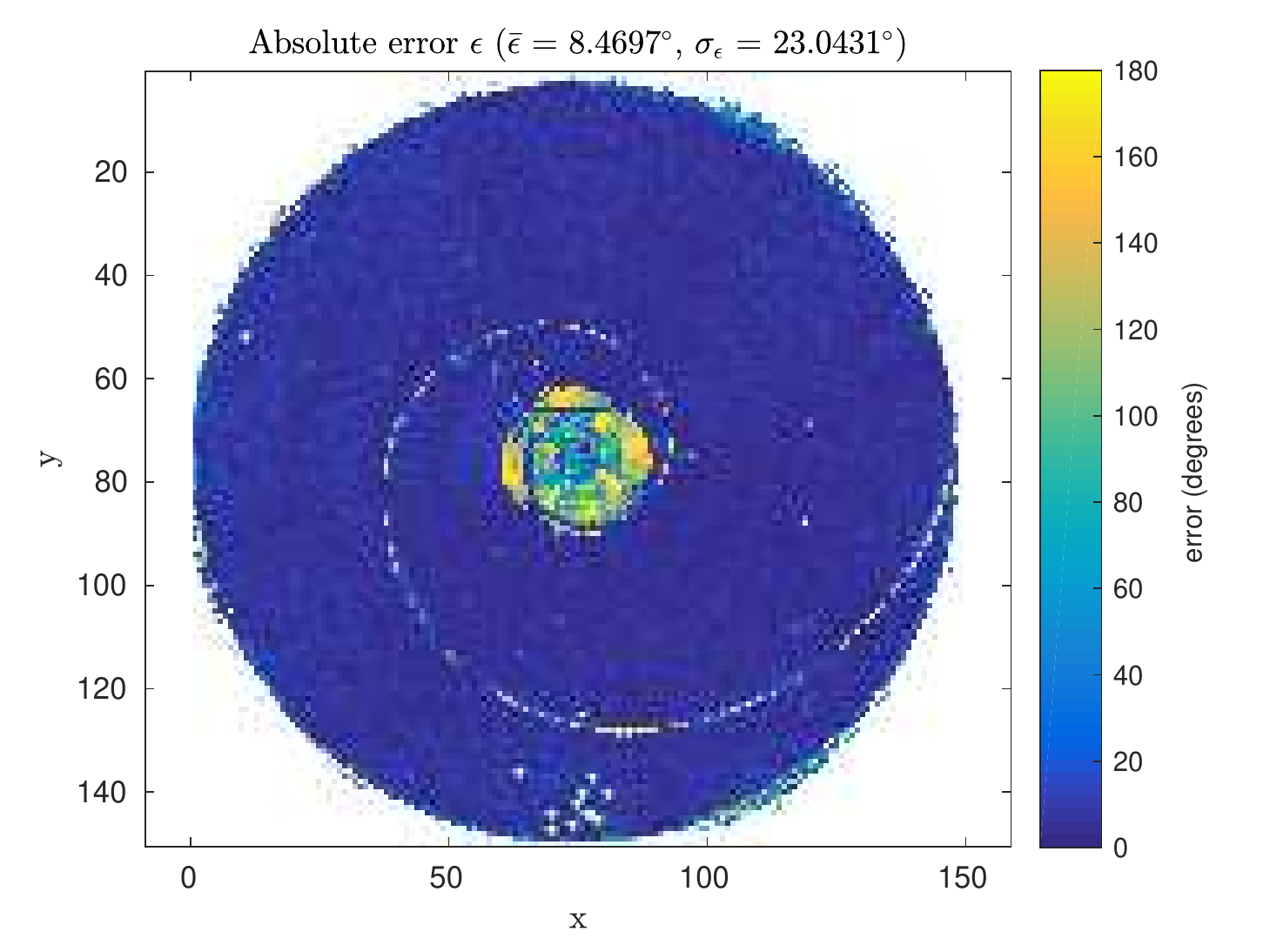}
\caption{}
\end{subfigure}
\begin{subfigure}[b]{0.49\textwidth}
\includegraphics[width=\textwidth]{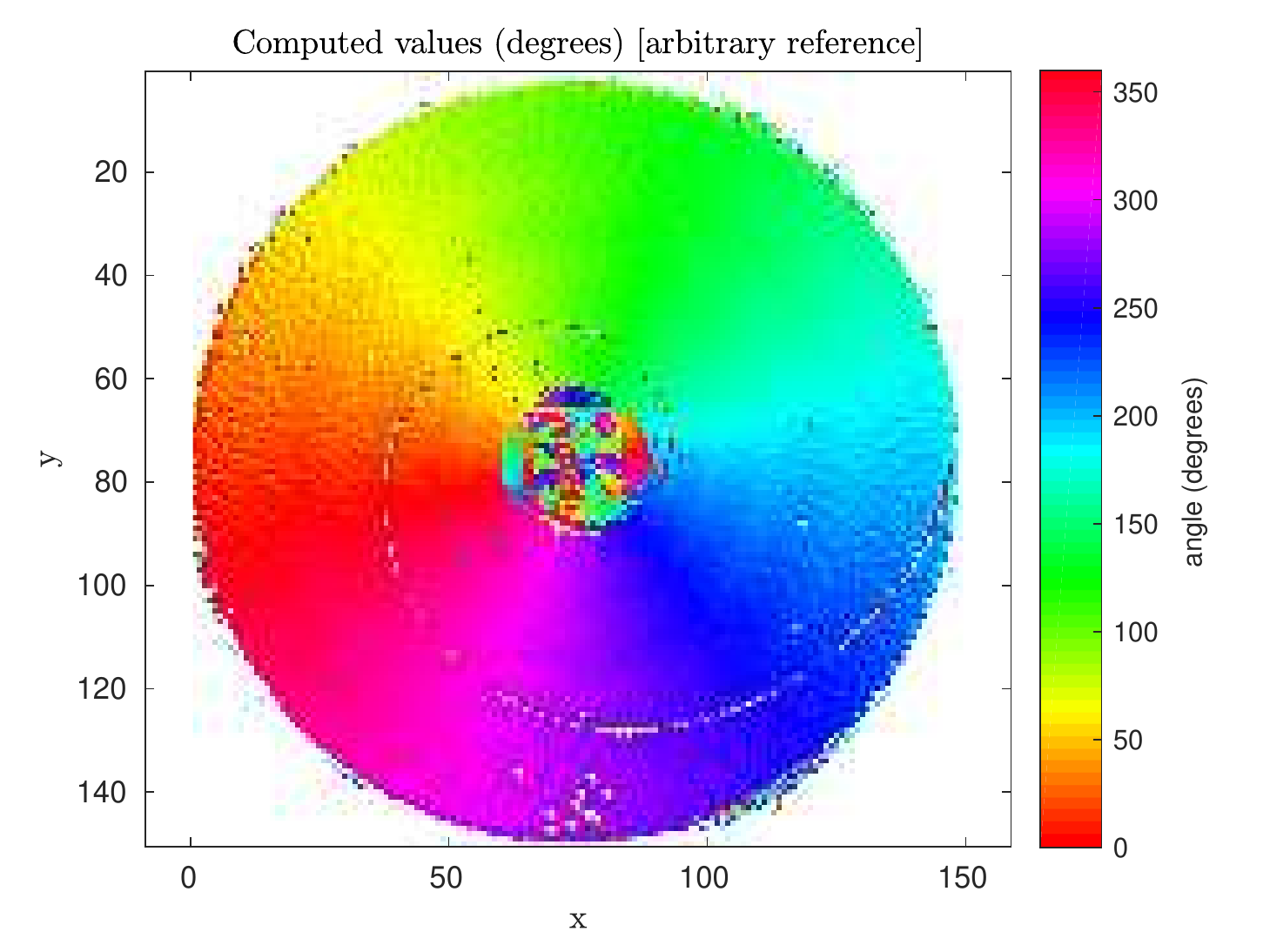}
\caption{}
\end{subfigure}
\caption{Rotating spiral. a) Absolute error. One can notice that the error is mainly located at the end of edges, due to the lack of adjacent active pixels. In some extend, it is impossible to compute a flow here. b) Top view of the output of the network.}
\label{fig::flow_outerror}
\end{figure}

The implemented network, as presented in Section~\ref{sec::implementation}, uses $\Delta X=6$ and $\Delta Y=6$, leading to a use of $2325$ cores, $558000$ neurons. Feeding this network with input from a rotating spiral, as shown in Fig.~\ref{fig:VisualSources}, gives results shown in Fig.~\ref{fig::flow_out1}. The estimated direction of movment is color-coded in the HSV space. Fig.~\ref{fig::flow_outerror} shows the estimated error for this network. Here, the proposed approach is able to extract the direction of movement with an absolute error mean of $\bar \epsilon \simeq 8.5\deg$. This error is mainly located at the end of the edges, where no neighborhood is available, thus given wrong measurements.

\begin{figure}
\centering
\includegraphics[width=0.49\textwidth]{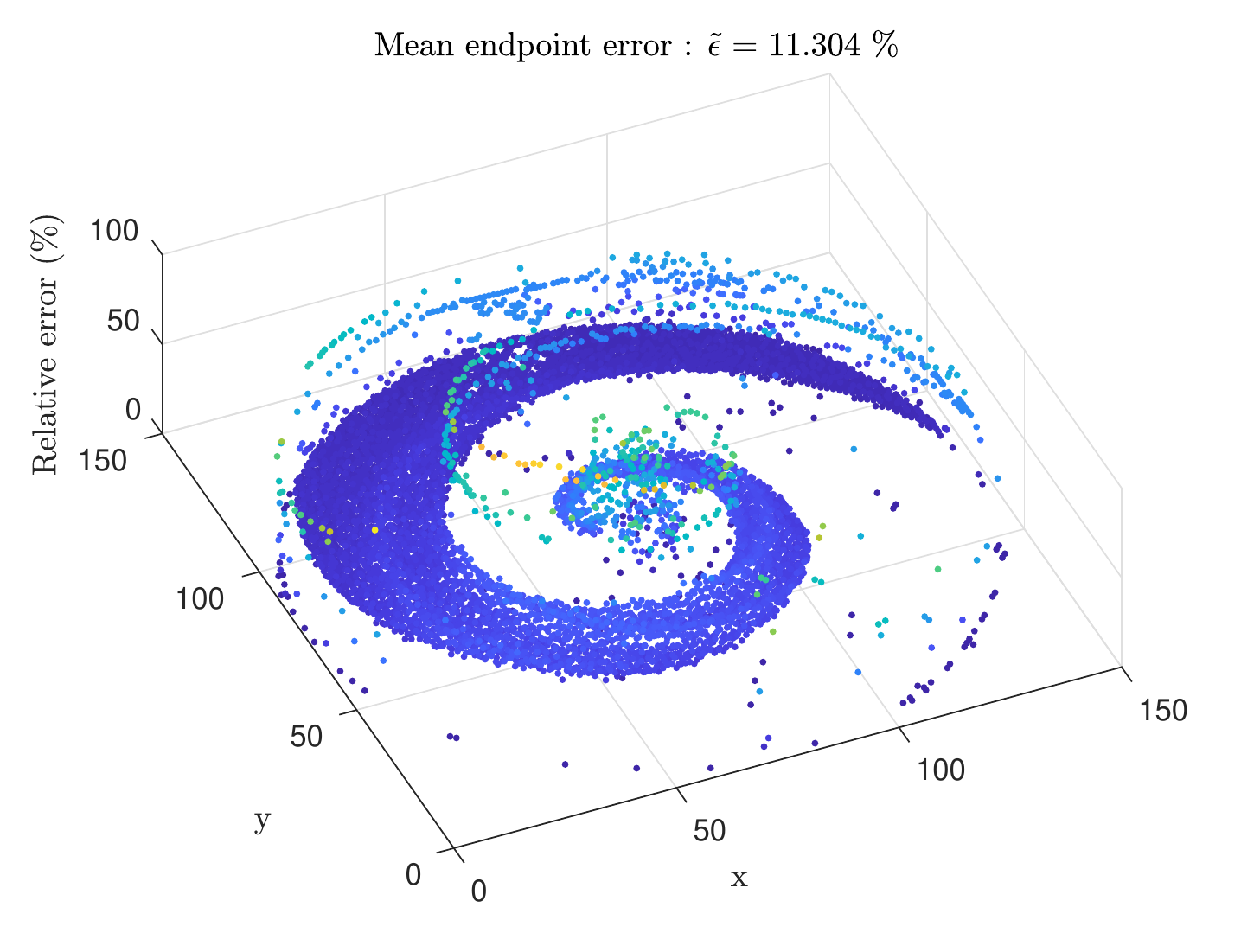}
\caption{Average Endpoint Error (AEE) for the spiral input. Our network is able to reconstruct the velocity vector with an average error of 11\%, error which is mainly located at the edges of the spiral, where it is impossible to now the speed by our method.}
\label{fig::AAE}
\end{figure}

Fig.~\ref{fig::flow_out1} shows a 3D representation of the optical flow estimates arising from the spiral data. Fig.~\ref{fig::flow_outerror} shows the angular error for the same spiral recording. Fig.~\ref{fig::AAE} shows the average endpoint error for the spiral data. A video of the spiral result is available online at \cite{spiral_video}.

\begin{figure}
\centering
\begin{subfigure}[b]{0.49\textwidth}
\includegraphics[width=\textwidth]{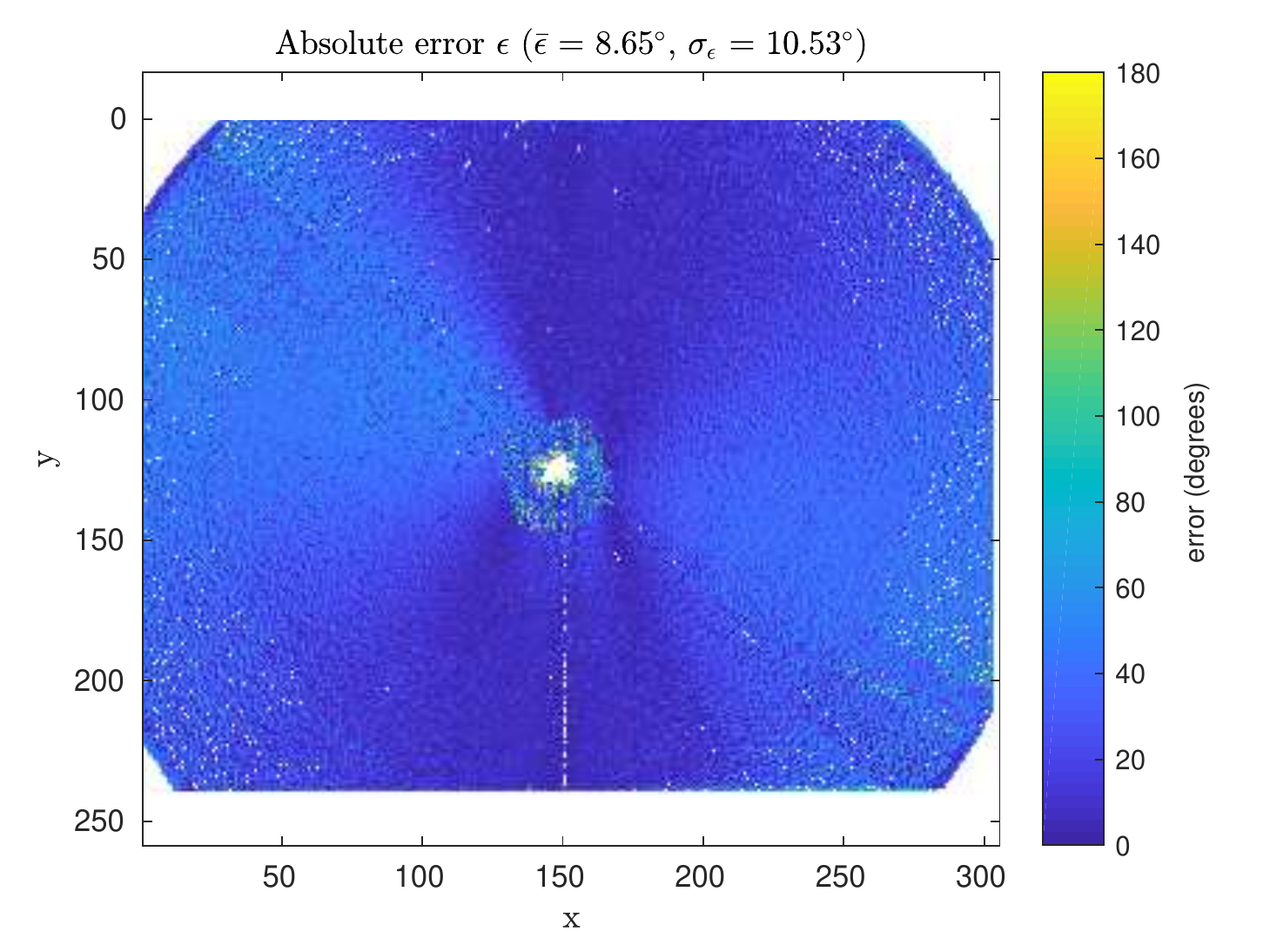}
\caption{}
\end{subfigure}
\begin{subfigure}[b]{0.49\textwidth}
\includegraphics[width=\textwidth]{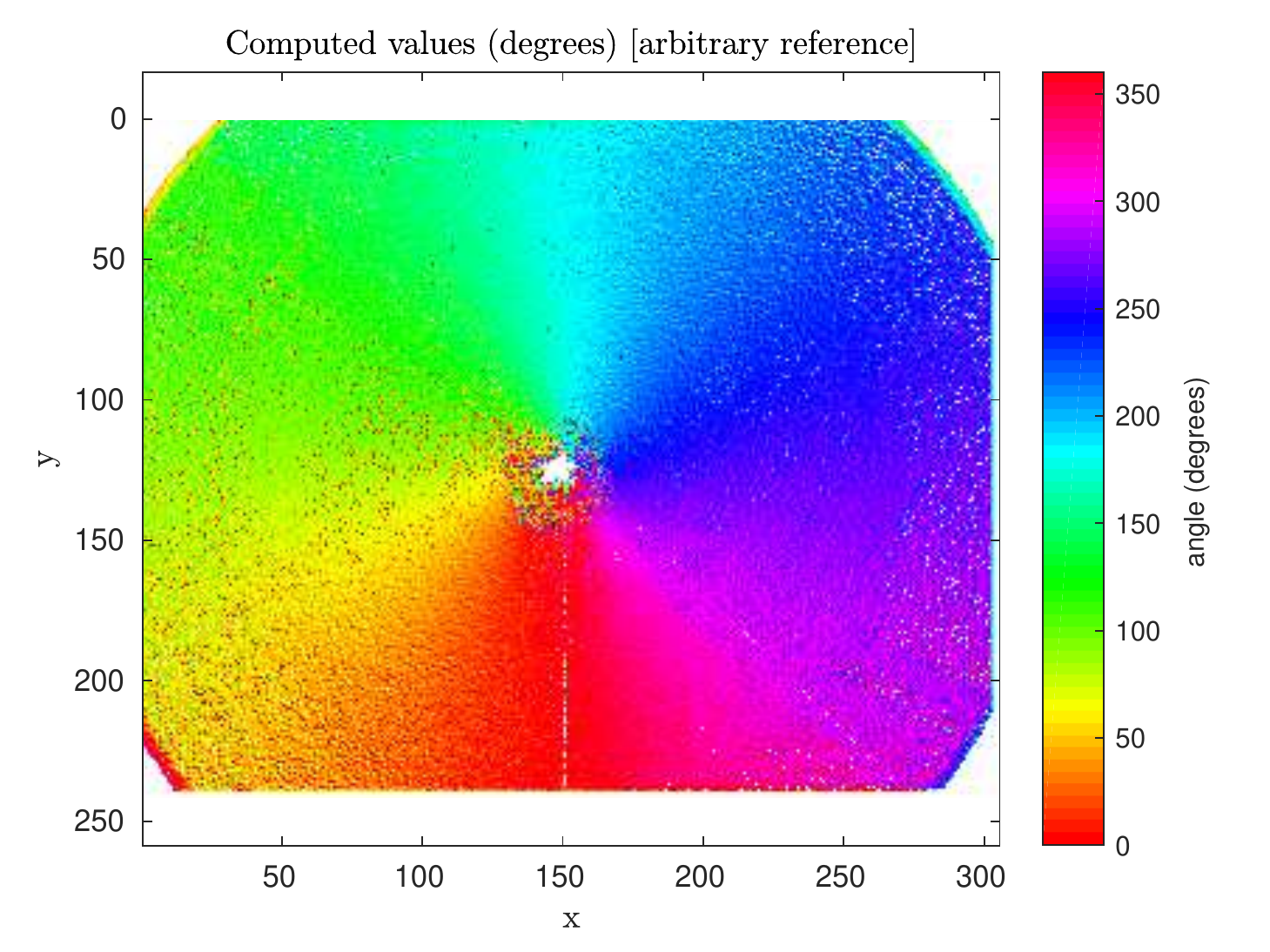}
\caption{}
\end{subfigure}
\caption{Rotating pipe. a) Absolute error. One can notice that the error is mainly located at the end of edges, due to the lack of adjacent active pixels. In some extend, it is impossible to compute a flow here. b) Top view of the output of the network. Video available online \cite{pipe_video}.}
\label{fig::flow_bar_outerror}
\end{figure}

Fig.~\ref{fig::flow_bar_outerror} shows similar angular error for the pipe data. A video of the pipe results can be found online at \cite{pipe_video}.

\begin{figure*}
\centering
\includegraphics[width=\textwidth]{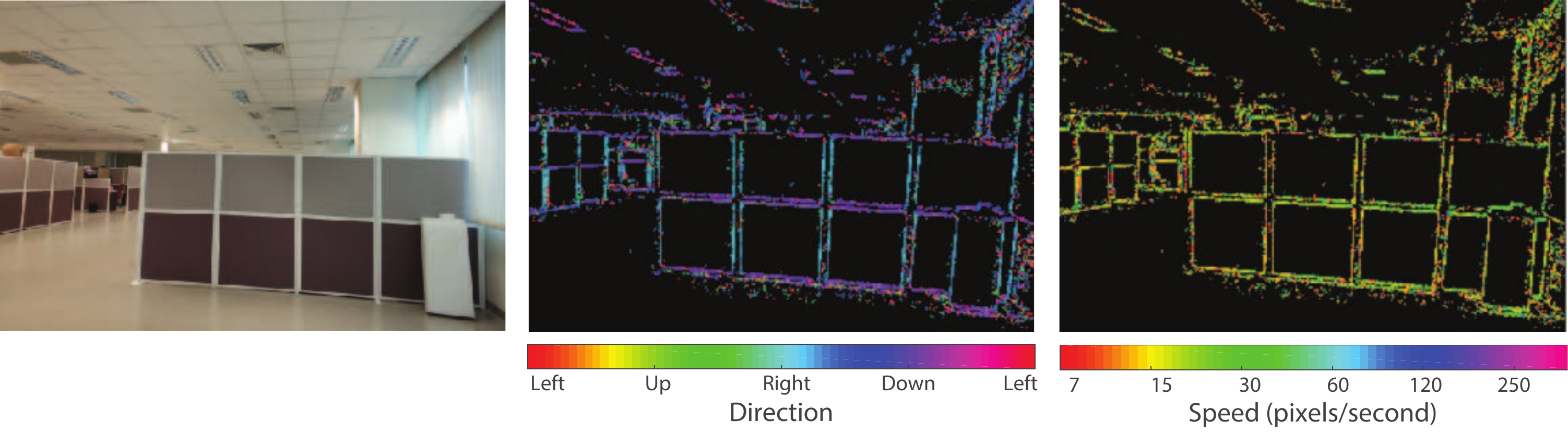}
\caption{Snapshot from a video showing optical flow output from TrueNorth for data captured while moving the ATIS sensor by hand in the lab. The leftmost image shows the scene from a similar angle captured using a cellphone camera. The middle shows the direction of motion detected, and the right shows the speed. Strong vertical and horizontal edges come out clearly, but the direction of motion differs because the network estimates the normal component of optical flow.}
\label{fig::lab_scene}
\end{figure*}

Fig.~\ref{fig::lab_scene} shows a snapshot of the output from data captured with the sensor hand held in the lab. A video of this data is available online at \cite{lab_scene_video}.

\section{Discussion}\label{sec:discussion}
The model presented here is capable of estimating visual motion in real-time with good accuracy, but there are some limitations.

The method described uses 57\% of the cores available on TrueNorth to estimate optical flow at every pixel over the full ATIS resolution of 304$\times$240. To extend the method to operate at higher resolutions, such as the prototype 640$\times$480 pixel DAVIS sensor, would require either using multiple TrueNorth chips, or only computing optical flow at every second pixel in the x and y directions.

Of the 256 neurons available on each TrueNorth core, 16 are unused. Of the 240 remaining neurons, 24 (10\%) are used to copy spike signals across the core boundaries to prevent motion estimation artifacts at the core boundary. Since copying the signal across the core boundary introduces a 1 tick delay, the inputs to each core must also be delayed 1 tick before processing, which uses another 6$\times$6=36 of the 240 used neurons. The block connectivity of TrueNorth has many advantages for parallelizing local computations in each core, but at the cost of requiring spikes to be copied between neighbouring blocks, which ends up using $36+16=52$ neurons per core.

Another constraint is imposed by the time resolution of TrueNorth, which is operating at a 1ms tick interval. This time quantization results in each DS unit's output speeds being quantized to $1/n, n\in \mathbb{N}$ pixels/ms. Such quantization can result in large error for large speeds (since the quantized bins are further apart at high speed), and speeds faster than 1 pixel/ms will be detected as 0.

The maximum detectable speed could be increased by placing DS input pixels further apart. A spacing of $k$ pixels between the DS unit input pixels would result in a maximum possible speed of $k$ pixels/ms. However, this approach would aggravate the edge correspondence problem, because it becomes less likely that the two pixels are seeing the same edge. Another possible method for increasing the maximum detectable speed is to reduce the tick period for TrueNorth. While this is definitely possible, a maximum speed of 1 pixel/ms was deemed good enough and reducing the tick period of TrueNorth was not explored further.

The slowest speed which can be detected is $1/\tau_r$, set by the length of the delay, $\tau_r$. Increasing the spacing of DS pixels to $k$ would also increase this minimum detectable speed to $k/\tau_r$.

There is a tradeoff between the slowest detectable stimulus, and the minimum allowable time between two subsequent stimuli. For two nearby, fast moving stimuli to each be uniquely detected and measured, their onsets must be spaced as least $\tau_r$ ms apart. Thus for fast moving stimuli it is desirable to have $\tau_r$ be small, but for slow stimuli, we require $\tau_r$ to be large.

Spike IO bandwidth of the TrueNorth chip also poses a constraint. The TrueNorth chip has four spike IO ports (North South East West), each capable of 40k spikes per millisecond. On the NS1e development platform used in this work, all outputs spikes are communicated to a Xilinx Zynq through one port. Although 40k spikes/ms is a lot, the model presented in this paper does generate a lot of output spikes. The worst case scenario happens in complex scenes with lots of slowly moving edges. Edge density increases the number of DS units which respond, while slow moving edges generate a lot more spikes than fast moving edges. An isolated noise event at a single pixel is a particularly bad culprit for generating spikes since it will simultaneously activate all four 4 DS units located at that pixel, and each of these DS units will generate a train of $\tau_r$ spikes before self-inhibiting.

If too many output spikes are generated, the TrueNorth chip tick period will extend to allow time for the spikes to be communicated, which will distort the estimated flow (which assumes a 1ms tick). In the case where pre-recorded data is being run through the model (instead of live ATIS data), the TrueNorth Neurosynaptic System will input spikes in accordance with the slowed down tick, thereby still allowing the model to estimate speeds accurately, but the system will run slower than real-time. On-chip spike communication is much faster than off-chip spike communication, so in future a method which interprets velocity on-chip and more efficiently communicates the results off-chip can help to mitigate the IO constraint.

The model described in this paper can be classified as a token based method, where a token is generated whenever an edge passes over a pixel. The actual gradient of the edge is never estimated, the gradient just has to be large enough to trigger a spike from the ATIS pixel. Large gradients may result in multiple spikes in response to a single edge, but the refractory period of the first stage will eliminate these secondary spikes. Much like many other event-based optical flow methods \cite{giulioni2016event, benosman2012asynchronous, barranco2014contour}, this method only estimates flow at edges, and it estimates normal flow (optical flow normal to the edge direction). However, some event-based methods do estimate gradients for use in optical flow computation \cite{barranco2014contour}, or use global constraints to estimate optical flow for image regions where little or no gradient is present \cite{bardow2016simultaneous}.

Currently the output spike signals must be interpreted off chip to extract the velocity information. Since 43\% of the TrueNorth cores are unused, there remains room to develop on-chip interpretation of the spikes to extract velocity. However, the output of the TrueNorth chip is always in spiking format, so such development work would simplify, but not eliminate, the off-chip velocity extraction.

\section{Conclusion}
A SNN method for normal optical flow computation from silicon retina data has been proposed. The method is capable of measuring the speed of edges in the range of 1/50 pixels/ms to 1 pixel/ms in real time. The light signal coming into the vision sensor is transduced into spikes by a silicon retina. These spikes are then fed into the SNN for processing, which provides an output estimate also encoded as spikes. This network, consuming $80mW$ ($70 mW$ for TrueNorth, $10mW$ for the ATIS, omitting FPGA used for communication, which can be removed for targetted applications), running real time and using 2325 cores, 558000 neurons, is able to extract optical flow with both a low angular error rate below $10$ degrees, and Average Enpoint Error (AEE \cite{baker2011database}) of 11\%, for a density estimation of 51\%.

\section*{Acknowledgments}
This work was partially achieved during the $22^{nd}$ edition of the Telluride neuromorphic workshop in Telluride, Colorado, 2016. The authors would like to thank the organizers and all the staff of this workshop for the fruitful discussions and exchanges that took place there.


\begin{IEEEbiography}
[{\includegraphics[width=1in,height=1.25in,clip,keepaspectratio]{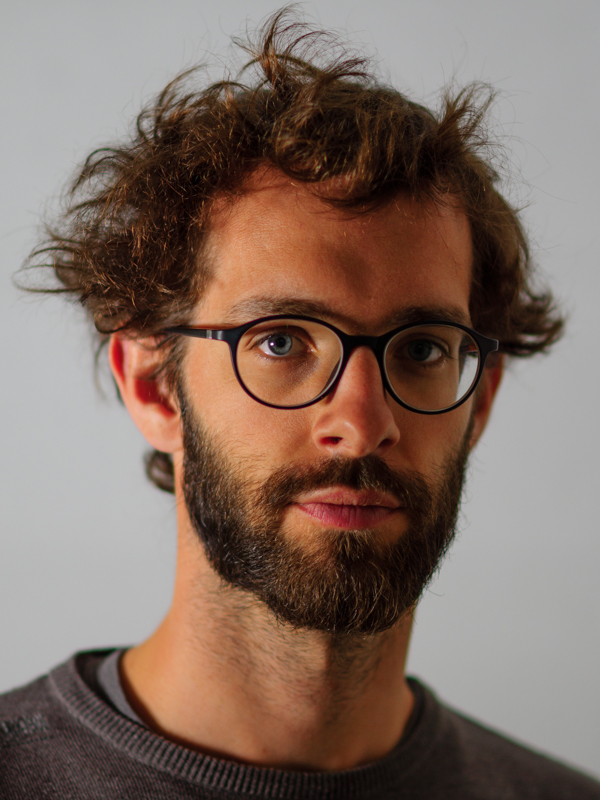}}]{Germain Haessig} received the B.Sc. and the Agrégation degree in electrical engineering from Ecole Normale Superieure de Cachan (2012, 2014), and the M.Sc degree in advanced systems and robotics from University Pierre and Marie Curie (2015). He is currently pursuing the Ph.D. degree in neurally inspired hardware systems and sensors within the Vision Institute, Paris.

\end{IEEEbiography}

\begin{IEEEbiography}
[{\includegraphics[width=1in,height=1.25in,clip,keepaspectratio]{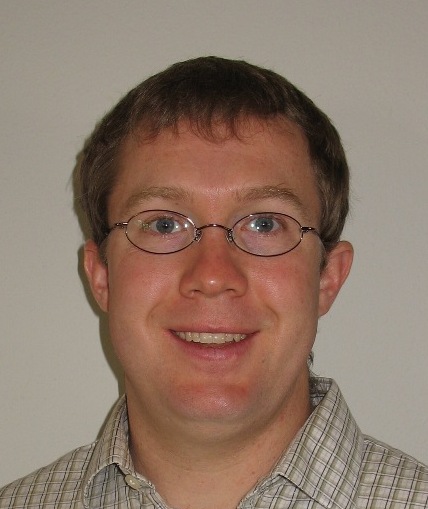}}]{Andrew Cassidy} (M’97) received the M.S. degree in electrical and computer engineering from Carnegie Mellon University, Pittsburgh, PA, USA, in 2003, and the Ph.D. degree in electrical and computer engineering from Johns Hopkins University, Baltimore, MD, USA, in 2010.

He is a Research Staff Member in the Brain-Inspired Computing group at IBM Research—Almaden, San Jose, CA, USA.  He was one of the lead researchers in developing the TrueNorth neurosynaptic processor, as a part of the DARPA SyNAPSE program.  His expertise is in non-traditional computer architecture, and he is experienced in architectural optimization, design, and implementation, particularly for machine learning algorithms and cognitive computing applications. He has authored over 30 publications in international journals and conferences.

\end{IEEEbiography}

\begin{IEEEbiography}
[{\includegraphics[width=1in,height=1.25in,clip,keepaspectratio]{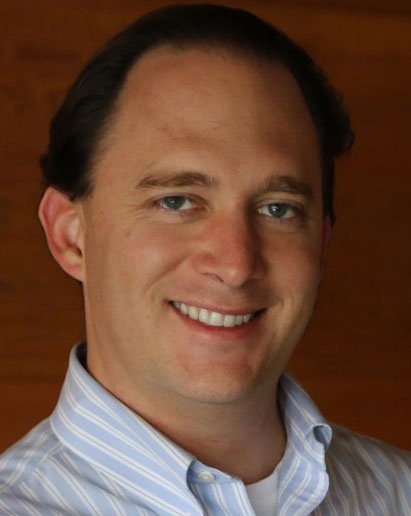}}]{Rodrigo Alvarez-Icaza} is a Research Staff Member in the Brain-Inspired Computing group at IBM Research Almaden where he was one of the lead developers of the TrueNorth processor. He has a Ph.D. degree in Bioengineering from Stanford University (2011) and a M.S. degree in Bioengineering from the University of Pennsylvania (2006). His research interests are focused on brain-inspired system architecture and their application on real-time control systems and robotics. He has authored over 40 patents and publications in international journals and conferences.
\end{IEEEbiography}

\begin{IEEEbiography}[{\includegraphics[width=1in,height=1.25in,clip,keepaspectratio]{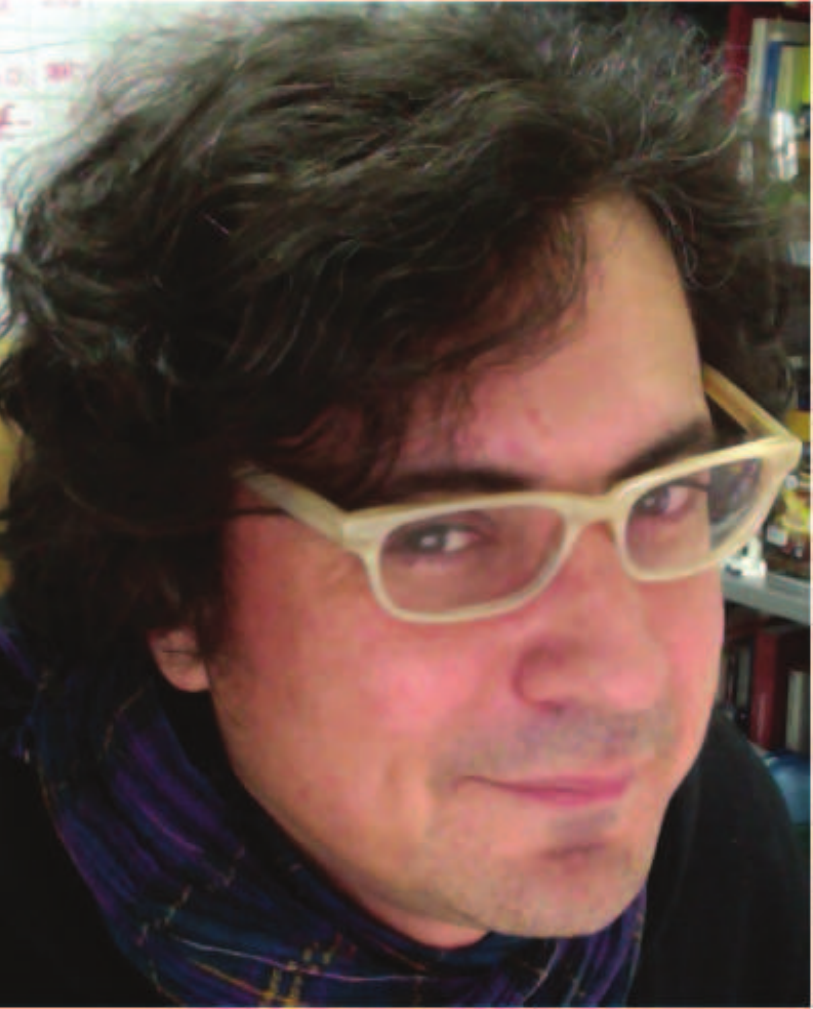}}]{Ryad Benosman} received the M.Sc. and Ph.D. degrees in applied mathematics and robotics from University Pierre and Marie Curie in 1994 and 1999, respectively. He is Associate Professor with University Pierre and Marie Curie, Paris, France, leading the Natural Computation and Neuromorphic Vision Laboratory, Vision Institute, Paris. His work covers neuromorphic visual computation and sensing. He is currently involved in the French retina prosthetics project and in the development of retina implants and cofounder of Pixium Vision a french prosthetics company. He is an expert in complex perception systems, which embraces the conception, design, and use of different vision sensors covering omnidirectional 360 degree wide-field of view cameras, variant scale sensors, and non-central sensors. He is among the pioneers of the domain of omni- directional vision and unusual cameras and still active in this domain. He has been involved in several national and European robotics projects, mainly in the design of artifcial visual loops and sensors. His current research interests include the understanding of the computation operated along the visual systems areas and establishing a link between computational and biological vision. Ryad Benosman has authored more than 100 scientific publications and holds several patents in the area of vision, robotics and image sensing. In 2013 he was awarded with the national best French scientific paper by the publication La Recherche for his work on neuromorphic retinas.
\end{IEEEbiography}

\begin{IEEEbiography}[{\includegraphics[width=1in,height=1.25in,clip,keepaspectratio]{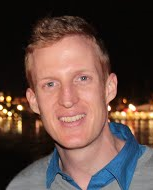}}]{Garrick Orchard} is a Senior Research Scientist at Temasek Laboratories and the Singapore Institute for Neurotechnology (SINAPSE) at the National University of Singapore. He holds a B.Sc. degree (with honours, 2006) in electrical engineering from the University of Cape Town, South Africa and M.S.E. (2009) and Ph.D. (2012) degrees in electrical and computer engineering from Johns Hopkins University, Baltimore, USA. He was named a Paul V. Renoff fellow in 2007, a Virginia and Edward M. Wysocki Sr. fellow in 2011, and a Temasek Research Fellow in 2015. He received the Johns Hopkins University Applied Physics Labs Hart Prize for Best Research and Development Project, and won the best live demonstration prize at the IEEE Biomedical Circuits and Systems conference 2012.  His research focuses on developing neuromorphic vision algorithms and systems for real-time sensing on mobile platforms. His other research interests include mixed-signal very large scale integration (VLSI) design, compressive sensing, spiking neural networks, visual perception, and legged locomotion.
\end{IEEEbiography}

\end{document}